# Women worry about family, men about the economy: Gender differences in emotional responses to COVID-19

Isabelle van der Vegt[1] and Bennett Kleinberg[1,2]

[1] Department of Security and Crime Science, University College London
[2] Dawes Centre for Future Crime, University College London
{isabelle.vandervegt, bennett.kleinberg}@ucl.ac.uk

**Abstract.** Among the critical challenges around the COVID-19 pandemic is dealing with the potentially detrimental effects on people's mental health. Designing appropriate interventions and identifying the concerns of those most at risk requires methods that can extract worries, concerns and emotional responses from text data. We examine gender differences and the effect of document length on worries about the ongoing COVID-19 situation. Our findings suggest that *i)* short texts do not offer as adequate insights into psychological processes as longer texts. We further find *ii)* marked gender differences in topics concerning emotional responses. Women worried more about their loved ones and severe health concerns while men were more occupied with effects on the economy and society. This paper adds to the understanding of general gender differences in language found elsewhere, and shows that the current unique circumstances likely amplified these effects. We close this paper with a call for more high-quality datasets due to the limitations of Tweet-sized data.

## 1 Introduction

The COVID-19 pandemic is having an enormous effect on the world, with alarming death tolls, strict social distancing measures, and far-reaching consequences for the global economy. In order to mitigate the impact that the virus may have on mental health, it is crucial to gain an understanding of the emotions, worries and concerns that the situation has brought about in people worldwide. Text data are rich sources to help with that task and computational methods potentially allow us to extract information about people's worries on a large scale. In previous work, the *COVID-19 Real World Worry Dataset* was introduced, consisting of 5,000 texts (2,500 short + 2,500 long) asking participants to express their emotions regarding the virus in written form [1]. In the current study, we delve deeper into that dataset and explore potential gender differences with regards to real-world worries about COVID-19. Building on a substantial evidence base for gender differences in language, we examine whether linguistic information can reveal whether and how men and women differ in how they respond to the crisis. Importantly, we also examine whether men or women are potentially more affected than the other, which may hold implications for developing mitigation strategies for those who need them the most.

## 1.1 Real World Worry Dataset

The Real World Worry Dataset (RWWD) is a text dataset collected from 2,500 people, where each participant was asked to report their emotions and worries regarding COVID-19 [1]. All participants were from the UK and completed the study through crowdsourcing platform Prolific Academic. The participants selected the most appropriate emotion describing what they were experiencing, choosing from anger, anxiety, desire, disgust, fear, happiness, relaxation, or sadness. They then rated the extent to which they worried about the COVID-19 situation and scored each of the eight emotions on a 9-point scale. Participants – all social media users – then wrote both a long text (avg. 128 tokens) and a short, Tweet-sized text (avg. 28 tokens) about said emotions and worries. The instructions read: "write in a few sentences how you feel about the Corona situation at this very moment. This text should express your feelings at this moment."

## 1.2 Gender differences in language

A large body of research has studied gender differences in language. Researchers have adopted both closed- and open vocabulary approaches. A closed vocabulary refers to approaches, where gender differences are measured through predefined word lists and categories. The LIWC (Linguistic Inquiry and Word Count) is a prominent example that measures linguistic categories (e.g., pronouns and verbs), psychological processes (e.g., anger and certainty), and personal concerns (e.g., family and money) [2]. The LIWC outputs the percentage of a document that belongs to each category. An open vocabulary approach is data-driven, in that gender differences are assessed without the use of predefined concepts or word lists. For example, $n$-grams or topics may be used to study gender differences.

In a closed-vocabulary study of 14,324 text samples from different sources (e.g. stream-of-consciousness essays, emotional writing), gender differences for LIWC categories were examined [3]. It was found that women used more pronouns (Cohen's $d = -0.36$)[1] words referring to emotions ($d = -0.11$), including anxiety ($d = -0.16$) and sadness ($d = -0.10$), as well as social words referring to friends ($d = -0.09$) and family ($d = -0.12$), and past ($d = -0.12$) and present-tense ($d = -0.18$) verbs. On the other hand, men used more articles ($d = 0.24$), numbers ($d = 0.15$), swear words ($d = 0.22$), and words referring to occupations ($d = 0.12$) and money ($d = 0.10$)[2]. Another approach partially replicated these results, showing that gender differences also emerged for language on social media (Facebook status updates from 75,000 individuals) [6]. In addition to examining differences in LIWC categories, the authors extended their approach to a data-driven, open-vocabulary approach, using both $n$-grams and topics. For example, unigrams such as 'excited', 'love', and 'wonderful' were used more frequently by women, whereas unigrams such as 'xbox', 'fuck', and 'government' were used more by men. In terms of topics, women more often mentioned family and friends and wished others a happy birthday, whereas men spoke more often about gaming and governmental/economic affairs [6].

---

[1] The effect size Cohen's $d$ expresses the magnitude of the difference after correcting for sample size and, therefore, offers a more nuanced metric than p-values [4]. A $d$ of 0.20, 0.50 and 0.80 can be interpreted as a small, moderate and large effect, respectively [5].

[2] Throughout this paper, positive values show that the respective concept is used more by men and negative values show that it is used more by women.

### 1.3 Contribution of this paper

With the current paper, we aim to add to the timely need to understand emotional reactions to the "Corona crisis". A step towards finding intervention strategies for individuals in need of mental health support is gaining an understanding of how different groups are affected by the situation. We examine the role of gender in the experienced emotions as well as potential manifestations of the emotional responses in texts. Specifically, we look at gender differences *i)* in self-reported emotions, *ii)* in the prevalence of topics using a correlated topic model approach [7], *iii)* between features derived from an open-vocabulary approach, and *iv)* in psycholinguistic constructs. We use the COVID-19 Real World Worry Dataset and thereby also test whether the differences emerge similarly in long compared to Tweet-sized texts.

## 2 Method

### 2.1 Data

The RWWD contains 5,000 texts of 2,500 individuals - each of whom wrote a long and short text expressing their emotions about COVID-19. We applied the same base exclusion criteria as [1] (i.e. nine participants who padded their statements with punctuation), and further excluded those participants who did not report their gender ($n = 55$). The sample used in this paper consisted of $n = 2,436$ participants, 65.15% female, 34.85% male.

### 2.2 Gender differences in emotion

We examined whether there were gender differences in the self-reported emotion scores (i.e. how people felt about COVID-19) using Bayesian hypothesis testing [8, 9]. In short, we report the Bayes factor, $BF_{10}$ which expresses the degree to which the data are more likely to occur under the hypothesis that there is a gender difference, compared to the null hypothesis (i.e. no gender difference). For example, $BF_{10} = 20$ means that the data are 20 times more likely under the hypothesis that there is a gender difference. Importantly, Bayesian hypothesis testing allows for a quantification of support for the null hypothesis, too. While a $BF_{10} = 1$ implies that the data are equally likely under both hypotheses, a $BF_{10}$ smaller than 1 indicates the support for the null, since $BF_{10} = \frac{1}{BF_{01}}$.

In addition, we report Bayesian credible intervals. The Bayesian credible interval is that interval in which the parameter of interest (here: the raw mean delta between female and male values) lies with a probability of 95% (i.e. it includes the 95% most credible values) [12, 13]. We calculated the equal-tailed credible interval with 10,000 sampling iterations.

### 2.3 Topic differences

As preprocessing steps, prior to topic modelling, all text data were lower-cased, stemmed, and all punctuation, stopwords, and numbers removed. First, we assess whether there are differences in topics between long and short texts. We construct a topic model[3] for all texts (long + short) and select the number of topics by examining values for semantic coherence and exclusivity [10, 11]. That approach assigns a probability for each document belonging to each topic. Here, we assign each document to its most likely topic (i.e., the highest topic probability for the document). We use a Chi-square test to examine whether there is an association between document type (long vs short) and topic occurrence. Standardised residuals (z-scores) are used to assess what drives a potential association.

Gender differences in topic occurrences are assessed in the same way for long and short texts separately. A Chi-square test is applied to test for an association between gender (male vs. female) and topic occurrence.

### 2.4 Open-vocabulary differences

In addition, we also look at differences in *n*-grams (unigrams, bigrams, trigrams) without the assumption of topic membership. Specifically, we calculate Bayes factors for male vs female comparisons on all *n*-grams to explore how both genders differ. We conduct that analysis for the short and long texts separately.

### 2.5 Closed-vocabulary differences

Since *n*-grams might not capture higher-order constructs (e.g., analytical thinking, anxiety) and psychological processes, we run the same analysis as in **2.4** using the LIWC 2015 [2].

## 3 Results

### 3.1 Gender differences in reported emotions

We compared the self-reported emotions (ranging from 1 = very low to 9 = very high) calculating Bayes factors and Bayesian credible intervals. Table 1 suggests that there was extreme evidence (based on the Bayes factor [8]) that women were more worried, anxious, afraid and sad than men. There was strong evidence that women were angrier than men. Conversely, men reported considerably more desire and more relaxation than women. We also assessed whether gender was associated with the "best fitting" chosen emotion.

A Chi-square test, $X^2(7) = 43.83$, $p < .001$, indicated an association between the chosen emotion and gender. Standardised residuals showed that this effect was driven by disparities between females choosing fear significantly more often than males ($z = -2.62$) and males choosing relaxation significantly more often ($z = 4.40$) than females.

---

[3] All topic models were constructed with the *stm* R package [10]

Thus, while anxiety was overall the most chosen emotion (55.36%, see [1]), gender did play a role for the selection of fear and relaxation.

Table 1. Means (SD) for reported emotions about COVID-19 and statistical test results for gender differences.

|  | Female | Male | $BF_{10}$ | Mean delta [95% Bayesian CI] |
|---|---|---|---|---|
| Worry | 6.84 (1.62) | 6.01 (1.90) | $> 10^{25}$ | -0.82 [-0.96; -0.68] |
| Anger | 4.04 (2.22) | 3.68 (2.26) | 54.97 | -0.36 [-0.54; -0.17] |
| Anxiety | 6.78 (2.14) | 5.95 (2.43) | $> 10^{14}$ | -0.82 [-1.01; -0.64] |
| Desire | 2.72 (1.96) | 3.48 (2.09) | $> 10^{15}$ | 0.76 [0.60; 0.94] |
| Disgust | 3.23 (2.13) | 3.21 (2.15) | 0.05 | -0.01 [-0.19; 0.17] |
| Fear | 6.00 (2.17) | 5.04 (2.34) | $> 10^{20}$ | -0.95 [-1.14; -0.76] |
| Happiness | 3.57 (1.87) | 3.74 (1.93) | 0.41 | 0.17 [0.01; 0.32] |
| Relaxation | 3.73 (2.10) | 4.36 (2.13) | $> 10^{9}$ | 0.62 [0.45; 0.79] |
| Sadness | 5.91 (2.21) | 4.96 (2.36) | $> 10^{19}$ | -0.95 [-1.13; -0.76] |

*Note.* Positive delta values indicate that the variable was scored higher by men than women, and vice versa.

### 3.2 Differences in topic occurrence

**Long vs short texts.**
For the topic model with long and short texts, 15 topics were selected based on semantic coherence and exclusivity of topic words. A significant association was found between text length (long vs. short) and topic occurrence, $X^2(14) = 1776.6$, $p < .001$. The six topics that differed most (i.e., highest standardised residuals) between long and short texts are depicted in Table 2[4]. Long texts were more likely to concern worries about both family and work, as well as the societal impact of the virus. Short texts were more likely to concern lockdown rules, staying home, and negative emotions.

Table 2. Topic differences (z-scores) long and short texts

| Topic | Terms | Std. residual |
|---|---|---|
| Lockdown rules | stay, home, follow, live, rule, peopl, nhs, help, wish, other, please, protect, listen, save, selfish | -22.53 |
| Family worries | worri, work, family, health, children, risk, catch, also, worker, job, food, elder, virus, shop, high | 19.00 |
| Staying home | can, keep, safe, stay, get, home, try, right, best, time, now, possible, position, make, love | -17.32 |
| Work worries | feel, also, lot, sad, work, quit, howev, family, find, job, make, anxious, other, difficult, anxiety | 13.93 |
| Negative emotions | feel, anxious, situat, scare, fear, sad, end, like, anxiety, come, whole, moment, corona, job, stress | -12.49 |
| Societal impact | concern, situat, health, impact, current, effect, economy, person, term, social, lockdown, mental, affect, may, society | 10.35 |

*Note.* A positive standardised residual indicates that the topic was more likely to occur in long texts.

---

[4] For a full list of topics and terms, see: https://osf.io/ktwhp/

**Gender differences.**

The topic model for long texts contained 20 topics[4]. For this model, we observed a significant association between gender and topic occurrence, $X^2(19) = 140.02$, $p < .001$. Table 3 show the topics with the largest gender difference and suggests that men were more likely to write about the national impact of the virus, worries about health, and the future outlook regarding COVID-19. Women spoke more about family and friends, as well as sickness and death.

We selected a model with 15 topics to represent the short texts. Here, we also observed a significant association between gender and topic occurrence, $X^2(14) = 101.47$ $p < .001$. Women spoke more about missing family and friends, and included more calls for solidarity. In contrast, men spoke more about the international impact of the virus, governmental policy, and returning to normal.

**Table 3.** Topic proportions with significant gender differences

| Topic | Terms | Std. residuals |
|---|---|---|
| **Long texts** | | |
| National impact | countri, govern, way, can, taken, get, feel, like, hope, world, also, response, much, better, china | 5.34 |
| Family and friends | feel, will, anxious, also, know, sad, famili, friend, worri, like, end, dont, see, day, situation | -4.45 |
| Sickness/death | worri, get, one, feel, also, sick, love, scare, anxious, die, husband, mum, fear, day, alone | -4.34 |
| Health worries | health, famili, worri, member, also, situat, concern, mental, able, risk, time, current, condition, older, however | 3.92 |
| Outlook | will, virus, year, people, world, death, get, also, know, catch, many, end, ill, feel, human | 3.43 |
| **Short texts** | | |
| Missing family & friends | famili, life, see, friend, like, day, really, miss, cant, enjoy, wait, week, even, can't, fed | -4.91 |
| International impact | world, covid-, concern, death, mani, covid, news, economy, real, never, test, must, infect, term | 4.82 |
| Policy | govern, scare, countri, social, distance, angri, proud, public, crisis, line, response, boris, frontline, fron, nhs | 4.18 |
| Solidarity | stay, home, everyone, live, safe, pleas, insid, save, protect, listen, nhs, indoor, advice, quicker, want | -3.79 |
| Return to normal | get, can, thing, normal, back, better, well, bit, possible, start, worse, hand, sick, told, tire | 3.38 |

*Note.* Positive standardised residuals indicate that the topic occurrence was higher for men than women, and vice versa.

7### 3.3 Open-vocabulary differences

Before extracting unigrams, bigrams and trigrams, the corpus was lower-cased and stopwords and punctuation were removed. Table 4 indicates that, in long texts, women use "anxious" (both in unigram and in bigram/trigram form) markedly more often than men, and mention "family" and "children" more often. The findings are, in part, corroborated for short texts, which, in addition, include the unigrams "scared" and "scary". Interestingly, unigrams which were more frequently used by men were "hopefully" and "calm". In broad lines, these findings reflect the differences found using a topic-based approach, where women expressed more fear and men were more likely to write about a (hopeful) return to normal. We also observe that *n*-gram-based differences are more pronounced in the longer texts than in shorter ones.

**Table 4.** Top 10 ngrams with the largest gender differences for long and short texts.

| *n*-gram | BF$_{10}$ | Mean delta [95% Bayesian CI] |
|---|---|---|
| | **Long texts** | |
| anxious | $> 10^{10}$ | -0.20 [-0.26; -0.15] |
| sad | $> 10^{9}$ | -0.16 [-0.20; -0.11] |
| feel | $> 10^{8}$ | -0.40 [-0.52; -0.29] |
| family | $> 10^{6}$ | -0.20 [-0.26; -0.13] |
| i_feel | $> 10^{6}$ | -0.29 [-0.38; -0.19] |
| makes | $> 10^{5}$ | -0.09 [-0.12; -0.06] |
| feel_very | $> 10^{4}$ | -0.08 [-0.10; -0.05] |
| children | $> 10^{4}$ | -0.09 [-0.12; -0.06] |
| feel_anxious | $> 10^{4}$ | -0.06 [-0.09; -0.04] |
| i_feel_anxious | $> 10^{4}$ | -0.05 [-0.08; -0.03] |
| | **Short texts** | |
| family | 308.00 | -0.05 [-0.08; -0.03] |
| sad | 201.16 | -0.04 [-0.06; -0.02] |
| scared | 66.41 | -0.04 [-0.06; -0.02] |
| friends | 14.19 | -0.03 [-0.05; -0.01] |
| scary | 8.03 | -0.02 [-0.03 -0.01] |
| hopefully | 7.47 | 0.02 [0.01; 0.03] |
| loved_ones | 7.41 | -0.02 [-0.04; -0.01] |
| calm | 7.19 | 0.01 [0.01; 0.02] |
| stay | 4.96 | -0.07 [-0.12; -0.03] |
| stay_home | 4.73 | -0.05 [-0.05 -0.01] |

*Note.* Negative delta values indicate that the variable was scored higher by women than men, and vice versa.

### 3.4 Closed-vocabulary differences

To capture potential differences in higher-order constructs, we also looked at gender differences for the LIWC variables. Table 5 suggests that men had a higher score on analytical thinking, used more articles and more "big" words (more than six letters).

Women, on the other hand, used more pronouns (all, personal pronouns, and first-person pronouns), more verbs, and expressed more anxiety and references to their family. We also observe that women had a substantially higher focus on the present than men.

For short texts, we see that the differences are less pronounced (BFs smaller than ten only constitute moderate evidence [8] and are ignored here). The data show that men scored higher on references to power and used more articles, while women used more conjunctions and scored higher on authenticity.

**Table 5.** Top 10 LIWC variables with the largest gender differences for long and short texts.

| LIWC variable | Meaning | $BF_{10}$ | Mean delta [95% Bayesian CI] |
| --- | --- | --- | --- |
| **Long texts** | | | |
| Analytic | Analytical thinking | $> 10^{24}$ | 10.59 [8.67; 12.51] |
| pronoun | Pronouns (all) | $> 10^{22}$ | -1.81 [-2.14; -1.48] |
| ppron | Personal pronouns | $> 10^{19}$ | -1.49 [-1.79; -1.20] |
| article | Articles | $> 10^{17}$ | 0.92 [0.73; 1.12] |
| verb | Verbs | $> 10^{14}$ | -1.36 [-1.67; -1.05] |
| anx | Emotional process: anxiety | $> 10^{13}$ | -0.63 [-0.78; -0.48] |
| i | First person singular | $> 10^{13}$ | -1.28 [-1.57; -0.98] |
| focuspresent | Time orientation: present | $> 10^{12}$ | -1.11 [-1.39; -0.84] |
| Sixltr | Words with more than 6 letters | $> 10^{10}$ | 1.44 [1.06; 1.82] |
| family | Social processes: family | $> 10^{8}$ | -0.33 [-0.42; -0.24] |
| **Short texts** | | | |
| power | Drives: power | 28.06 | 0.64 [0.28; 0.98] |
| conj | Conjunctions | 24.66 | -0.84 [-1.30; -0.37] |
| Authentic | Authenticity | 23.01 | -5.48 [-8.69; -2.46] |
| article | Articles (all) | 12.22 | 0.72 [0.29; 1.13] |
| money | Personal concerns: money | 8.28 | 0.21 [0.08; 0.33] |
| work | Personal concerns: work | 3.91 | 0.37 [0.13; 0.62] |
| Analytic | Analytical thinking | 3.32 | 4.08 [1.36; 6.79] |
| home | Personal concerns: home | 3.27 | -0.44 [-0.73; -0.14] |
| i | First person singular | 2.61 | -0.64 [-1.08; -0.19] |
| insight | Cognitive processes: insight | 1.73 | -0.39 [-0.68; -0.11] |

*Note.* Positive delta values indicate that the variable was scored higher by men than women, and vice versa.

### 3.5 Follow-up exploration: concerns, drives and time focus.

To understand gender differences in emotional responses to COVID-19 on the psycholinguistic level better, we zoom in on three LIWC clusters (Table 6). We look at the clusters "personal concerns", "drives", and "time orientation" - each of which consists of sub-categories (e.g., concerns: work, death, drives: risk, achievement, time orientation: future, present). Men scored higher on the variables risk (e.g., cautious, dangerous), work, and money, whereas women had higher values for affiliation (e.g., friend, party, relatives), home (e.g., neighbour, pet) and a focus on the present. Again, these bottom-up findings seem to align with the topic models from a psycholinguistic angle.

**Table 6.** Means (SD) of drives, concerns and time orientation LIWC clusters and test results for gender differences.

| Variable | Females | Males | $BF_{10}$ | Mean delta [95% Bayesian CI] |
|---|---|---|---|---|
| Drives: affiliation | 2.99 (2.04) | 2.61 (2.09) | 594.41 | -0.38 [-0.55; -0.21] |
| Drives: achievement | 1.73 (1.32) | 1.74 (1.32) | 0.05 | 0.01 [-0.10; 0.12] |
| Drives: power | 2.30 (1.61) | 2.35 (1.75) | 0.06 | 0.05 [-0.09; 0.18] |
| Drives: reward | 1.33 (1.17) | 1.28 (1.14) | 0.07 | 0.05 [-0.05; 0.14] |
| Drives: risk | 0.97 (0.98) | 1.22 (1.18) | $> 10^3$ | 0.25 [0.16; 0.34] |
| Concerns: work | 2.24 (1.76) | 2.68 (2.02) | $> 10^3$ | 0.44 [0.28; 0.59] |
| Concerns: leisure | 1.18 (1.15) | 1.14 (1.22) | 0.07 | -0.04 [-0.14; 0.06] |
| Concerns: home | 1.24 (1.11) | 0.98 (1.07) | $> 10^3$ | -0.26 [-0.35; -0.17] |
| Concerns: money | 0.73 (0.98) | 0.96 (1.17) | $> 10^3$ | 0.22 [0.13; 0.31] |
| Concerns: work | 0.07 (0.31) | 0.07 (0.26) | 0.05 | 0.00 [-0.03; 0.02] |
| Concerns: death | 0.31 (0.56) | 0.29 (0.61) | 0.06 | -0.02 [-0.06; 0.03] |
| Time orientation: past | 1.80 (1.68) | 1.85 (1.73) | 0.07 | 0.06 [-0.08; 0.20] |
| Time orientation: present | 15.10 (3.23) | 13.98 (3.50) | $> 10^{11}$ | -1.11 [-1.40; -0.84] |
| Time orientation: future | 2.22 (1.68) | 2.32 (1.83) | 0.12 | 0.10 [-0.04; 0.24] |

*Note.* Positive delta values indicate that the variable was scored higher by men than women, and vice versa.

## 4   Discussion

This study elucidated gender differences in emotional responses to COVID-19 in several (linguistic) domains. Analyses were performed for both long and short, Tweet-sized texts. We first discuss the implications of using the respective text lengths for understanding emotions and worries. Then, we review the observed gender differences and relate them to previous research on this topic. Lastly, some potential future avenues for understanding worries from text data are identified.

### 4.1   Short vs long texts

In previous work using the same dataset, it was suggested that important differences emerge when participants are asked to write about their worries in long versus short, Tweet-size texts: in topic models that were constructed for long and short texts separately, longer texts seemed to shed more light on the worries and concerns of participants. In contrast, shorter texts seemed to function as a call for solidarity [1]. In the current study, we were able to statistically test that idea by constructing a topic model for both text types together. Indeed, when testing for an association between text length and topic occurrence, we found that topics significantly differed between text types. Our results confirmed that short Tweet-like texts more frequently referred to calls to 'stay at home, protect the NHS, save lives' (a UK government slogan during the crisis at the time of data collection). Longer texts more effectively related to the specific worries participants had, including those about their family, work, and society.

### 4.2 A cautionary note on using short text data

The apparent differences between long and Tweet-sized texts suggest that researchers need to exercise caution in relying largely on Twitter datasets to study the COVID-19 crisis, and other more general social phenomena. Indeed, several Twitter datasets have been released [12–16] for the research community to study responses to the Corona crisis. However, the current study shows that such data may be less useful if we are interested in gaining a deep understanding of emotional responses and worries. The exact reasons for that difference (e.g., possibly different functions that each text form serves) merit further research attention. One possibility is that instructing participants to write texts of a specific length also influences emotion expression. However, the observation that Tweet-sized texts in this study are less able than long texts to convey emotions and worries about COVID-19, on top of the classical limitations of social media data (e.g., demographic biases [17], participation biases [18], or data collection biases [19]), are reasons to be more cautious for issues as timely and urgent as mental health during a global pandemic. Ultimately, making use of the convenience of readily-available social media data might come at the expense of capturing people's concerns and could lead to skewed perceptions of these worries. We urge us as a research community to (re-) consider the importance of gathering high-quality, rich text data.

### 4.3 Emotional responses to COVID-19: males vs females

Gender differences emerged throughout this paper. Reported emotions showed that women were more worried, anxious, afraid, and sadder then men and these results were supported by linguistic differences. For instance, topic models suggested that women discussed sickness and death more often than men. *N*-gram differences showed that women used 'anxious', 'sad', and 'scared' more than men and LIWC differences showed that women used more words related to anxiety. Importantly, this is not to say that men did not worry about the crisis, as reported negative emotions were still relatively high (e.g., the average score of 6 out of 9 for worrying). Furthermore, topic models showed that men wrote more frequently about worries related to their health than women.

The results further illustrated differences in what men and women worry about with regards to the crisis. Women's focus on family and friends emerged consistently in the topic models, *n*-gram differences, and LIWC differences. On the other hand, men more frequently worried about national and international impact (topic models) and wrote more frequently about money and work than women (LIWC). All in all, these results seem to follow previous work on gender differences in language more generally. For example, similar to our results, previous studies have found that women, in general, score higher than men on the LIWC categories for anxiety, family, and home [3]. Our results also seemed to have further replicated that men use more articles, and use the LIWC categories money and work more often than women [3]. In light of these findings, it is of key importance to discern whether the gender differences that emerged in this study are specific to COVID-19 worries, or are simply a reflection of gender differences that emerge regardless of the issue in question.

There are some indications in our data to imply that the differences are in line with general gender differences but in a more pronounced way. Although there is no agreed upon (meta-analytic) benchmark of gender differences in language to which we can compare our results, there is some previous work with which comparisons can be made [3]. For example, general linguistic differences in anxiety ($d = -0.16$, here: $d = -0.36$), family ($d = -0.12$, here: $d = -0.34$), present focus ($d = -0.18$, here: $d = -0.33$), article use ($d = 0.24$, here: $d = 0.38$), references to work ($d = 0.12$, here: $d = 0.23$) and money ($d = 0.10$, here: $d = 0.20$) have been found previously [3]. All of these are present in our data as well but often with an effect twice the size (note: the $d$ values are calculated for comparison purpose but are not included in the tables). Thus, it is possible that the COVID-19 situation and the accompanying toll on mental health exacerbated the language differences between men and women further. If we follow the line of previous work [3], the intensified linguistic gender differences can be interpreted as serving different functions for men and women. It has been suggested that women predominantly discuss other people and internal, emotional processes, while men focus more on external events and objects [3, 6]. Similar patterns are discoverable in our data, where women were more likely to discuss loved ones and their own negative emotions, whereas men were more likely to write about the external effects of the virus on society. It is important to note that the findings here (and elsewhere) do not imply that these patterns are all-encompassing or exclusive to men or women: men do also discuss their family and women also worry about the economy, but on average the reported differences emerged. All in all, the current results seem to fall in line with previous empirical work as well as persisting gender stereotypes.

### 4.4    Next steps in understanding worries from text data

For the social and behavioural sciences during and after the Corona crisis, a principal research question revolves around the mental health implications of COVID-19. The current study leveraged text data to gain an understanding of what men and women worry about. At the same time, it is of vital interest to develop mitigation strategies for those who need them the most. While this paper attempted to shed light on gender differences, other fundamental questions still need answering.

First, relatively little is known about the manifestation of emotions in language and the subsequent communication of it in the form of text data (e.g., how good are people in expressing their emotions and, importantly, which emotions are better captured computationally than others?). Second, the type of a text (e.g., stream-of-consciousness essay vs pointed topical text vs Tweet) seems to determine the findings to a great deal (i.e. different topics, different effect sizes). Ideally, future research can illuminate how the language and aims of an individual change depending on the type of text. Third, to map out the worries on a large scale and use measurement studies to understand the concerns people have, further data collection in different (offline) contexts and participant samples beyond the UK alone will be needed. Sample demographics (e.g. socio-economic status, ethnicity) in this study could have influenced text writing or emotion expression, thus replications in different samples will be needed to understand these potential effects. Further large-scale data collection will also allow for increased attention to prediction models constructed on high-quality ground truth data.

## 5 Conclusion

The current paper contributes to our understanding of gender differences in worries related to the COVID-19 pandemic. Gender was related to the reported emotions, topics, *n*-grams, and psycholinguistic constructs. Women worried mainly about loved ones and expressed negative emotions such as anxiety and fear, whereas men were more concerned about the broader societal impacts of the virus. The results showed that long texts differed from short Tweet-size texts in reflecting emotions and worries.

**References**

1. Kleinberg, B., van der Vegt, I., & Mozes, M. (2020). Measuring Emotions in the COVID-19 Real World Worry Dataset. *arXiv:2004.04225 [cs]*. Retrieved from http://arxiv.org/abs/2004.04225

2. Pennebaker, J. W., Boyd, R. L., Jordan, K., & Blackburn, K. (2015). The Development and Psychometric Properties of LIWC2015. The University of Texas at Austin. Retrieved from https://repositories.lib.utexas.edu/handle/2152/31333

3. Newman, M. L., Groom, C. J., Handelman, L. D., & Pennebaker, J. W. (2008). Gender Differences in Language Use: An Analysis of 14,000 Text Samples. *Discourse Processes*, *45*(3), 211–236. https://doi.org/10.1080/01638530802073712

4. Lakens, D. (2013). Calculating and reporting effect sizes to facilitate cumulative science: a practical primer for t-tests and ANOVAs. *Frontiers in Psychology*, *4*. https://doi.org/10.3389/fpsyg.2013.00863

5. Cohen, J. (1988). *Statistical Power Analysis for the Behavioral Sciences*. Academic Press.

6. Schwartz, H. A., Eichstaedt, J. C., Kern, M. L., Dziurzynski, L., Ramones, S. M., Agrawal, M., … Ungar, L. H. (2013). Personality, Gender, and Age in the Language of Social Media: The Open-Vocabulary Approach. *PLoS ONE*, *8*(9), e73791. https://doi.org/10.1371/journal.pone.0073791

7. Blei, D. M., & Lafferty, J. D. (2007). A correlated topic model of Science. *Annals of Applied Statistics*. Retrieved from https://projecteuclid.org/euclid.aoas/1183143727

8. Ortega, A., & Navarrete, G. (2017). Bayesian Hypothesis Testing: An Alternative to Null Hypothesis Significance Testing (NHST) in Psychology and Social Sciences. *Bayesian Inference*. https://doi.org/10.5772/intechopen.70230

9. Wagenmakers, E.-J., Lodewyckx, T., Kuriyal, H., & Grasman, R. (2010). Bayesian hypothesis testing for psychologists: A tutorial on the Savage–Dickey method. *Cognitive Psychology*, *60*(3), 158–189. https://doi.org/10.1016/j.cogpsych.2009.12.001

10. Roberts, M. E., Stewart, B. M., & Tingley, D. (2014). stm: R Package for Structural Topic Models. *Journal of Statistical Software*, 41.

11. Mimno, D., Wallach, H., Talley, E., Leenders, M., & McCallum, A. (2011). Optimizing Semantic Coherence in Topic Models, 11.

12. Kruschke, J. K. (2013). Bayesian estimation supersedes the t test. *Journal of Experimental Psychology: General*, *142*(2), 573–603. https://doi.org/10.1037/a0029146

13. Kruschke, J. K., & Liddell, T. M. (2018). The Bayesian New Statistics: Hypothesis testing, estimation, meta-analysis, and power analysis from a Bayesian perspective. *Psychonomic Bulletin & Review*, *25*(1), 178–206. https://doi.org/10.3758/s13423-016-1221-4

14. Banda, J. M., Tekumalla, R., Wang, G., Yu, J., Liu, T., Ding, Y., & Chowell, G. (2020, April 5). A Twitter Dataset of 150+ million tweets related to COVID-19 for open research. Zenodo. https://doi.org/10.5281/zenodo.3738018

15. Chen, E., Lerman, K., & Ferrara, E. (2020). *#COVID-19: The First Public Coronavirus Twitter Dataset*. Python. Retrieved from https://github.com/echen102/COVID-19-TweetIDs

16. Lamsal, R. (2020, March 13). Corona Virus (COVID-19) Tweets Dataset. IEEE. Retrieved from https://ieee-dataport.org/open-access/corona-virus-covid-19-tweets-dataset

17. Jacobs, C. (2020). *Coronada: Tweets about COVID-19*. Python. Retrieved from https://github.com/BayesForDays/coronada

18. Basile, V., & Caselli, T. (2020). TWITA - Long-term Social Media Collection at the University of Turin. Retrieved April 17, 2020, from http://twita.di.unito.it/dataset/40wita

19. Morstatter, F., Pfeffer, J., Liu, H., & Carley, K. M. (2013). Is the Sample Good Enough? Comparing Data from Twitter's Streaming API with Twitter's Firehose. *arXiv:1306.5204 [physics]*. Retrieved from http://arxiv.org/abs/1306.5204

20. Solymosi, R., Bowers, K. J., & Fujiyama, T. (2018). Crowdsourcing Subjective Perceptions of Neighbourhood Disorder: Interpreting Bias in Open Data. *The British Journal of Criminology*, *58*(4), 944–967. https://doi.org/10.1093/bjc/azx048

21. Pfeffer, J., Mayer, K., & Morstatter, F. (2018). Tampering with Twitter's Sample API. *EPJ Data Science*, *7*(1), 50. https://doi.org/10.1140/epjds/s13688-018-0178-0